\documentclass{article} 
\usepackage{iclr2027_conference,times}

\usepackage{graphicx}
\usepackage{booktabs}
\usepackage{pifont}   
\usepackage{amsmath}
\usepackage{amssymb}
\usepackage{hyperref}
\usepackage{url}

\usepackage{xcolor}
\usepackage{framed}
\usepackage{listings}
\definecolor{promptbg}{RGB}{244,246,250}
\definecolor{promptrule}{RGB}{110,140,176}
\definecolor{trajbg}{RGB}{247,247,244}
\definecolor{trajrule}{RGB}{150,150,150}
\newenvironment{promptbox}[1]{%
  \par\smallskip\noindent{\footnotesize\bfseries\sffamily #1}\par\nobreak\vskip1pt%
  \setlength{\FrameSep}{7pt}\setlength{\FrameRule}{2.5pt}%
  \MakeFramed{\hsize=\dimexpr\hsize-2\FrameSep-2.5pt\relax\FrameRestore}\itshape\small}%
{\endMakeFramed\par\smallskip}
\lstdefinestyle{traj}{basicstyle=\ttfamily\scriptsize,breaklines=true,
  backgroundcolor=\color{trajbg},frame=leftline,rulecolor=\color{trajrule},
  framerule=1.5pt,xleftmargin=6pt,framexleftmargin=6pt,aboveskip=6pt,belowskip=4pt,
  columns=fullflexible,keepspaces=true}

\newcommand{\nRecordings}{247}       
\newcommand{\nInstances}{390}        
\newcommand{\nHours}{2{,}536}        
\newcommand{\nAFcore}{214}           

\title{ECG-Scroll: A Long-Horizon, Streaming Benchmark and Agent Environment\\
for Interpretation of Ambulatory Electrocardiograms}

\iclrfinalcopy
\author{Haitao Li$^{1,2}$\quad Chenglin Li$^{1,2}$\quad Zhengyao Ding$^{1}$\quad Ziyu Li$^{1}$\quad Yiheng Mao$^{1}$\quad Zhengxing Huang$^{1}$\\[2pt]
{\normalfont $^{1}$Zhejiang University\quad $^{2}$Shanghai Innovation Institute}}

\begin{document}
\maketitle
\thispagestyle{fancy}\lhead{arXiv preprint}\rhead{}\chead{}  
\pagestyle{fancy}\fancyhead{}\lhead{arXiv preprint}

\begin{abstract}
Multimodal large language models (MLLMs) can now interpret a standard ten-second, twelve-lead electrocardiogram (ECG) with clinically grounded, reward-verified reasoning. Real cardiac monitoring is different. Ambulatory (Holter) and telemetry recordings span hours to days and are read as they stream in, and their clinically decisive findings are \emph{paroxysmal}, brief episodes buried in an otherwise unremarkable trace. Such a recording cannot be held in one context at diagnostic resolution, and its future has not yet happened, so a reader must work \emph{online}, deciding what to measure now, committing evidence to memory as it passes, and reporting events as they occur. We recast long-duration ECG interpretation as a long-horizon, \textbf{online (streaming, causal)} sequential decision process and introduce \textbf{ECG-Scroll}. As a benchmark, long ambulatory recordings are streamed to an agent chunk by chunk, and it must localize, quantify, and promptly flag paroxysmal events without access to future signal; because the underlying signal is retained, every answer is checkable against objective ground truth, giving rule-based rather than judge-based rewards, and the streaming formulation adds a metric batch evaluation cannot express, the \emph{detection latency} between an event's onset and the moment the agent records it. As an agent environment, it is a fixed, gym-style interaction layer that exercises three competencies single-glance ECG models never touch: \textbf{Memory}, \textbf{Tool} use through signal-grounded measurement rather than reading pixels, and \textbf{Planning} of what to measure now and when to commit. We release \nInstances{} whole-recording instances spanning \nHours{} hours of two-lead ambulatory ECG and evaluate a signal-threshold rule agent alongside off-the-shelf LLM agents online, characterizing how they use memory, tools, and planning and where the benchmark's head-room lies. Code and data: \url{https://github.com/CuCl-2/ECG-Scroll/}.
\end{abstract}


\section{Introduction}
Multimodal large language models (MLLMs) have advanced electrocardiogram (ECG) interpretation with remarkable speed, part of a broader wave of deep learning for cardiac signals that spans cross-modal generation and disease prediction~\citep{ding2026generating,ding2025ai}. Recent systems align ECG signals or ECG images with clinical text and produce diagnoses grounded in measurable waveform evidence. GEM~\citep{gem2025} couples a time-series encoder with a twelve-lead ECG image and ties each conclusion to quantitative parameters such as the QRS and PR intervals. ECG-R1~\citep{ecgr12026} generates protocol-guided instruction data and trains with reinforcement learning under diagnostic evidence rewards, yielding interpretations that are both reasoned and reward-verified. PULSE~\citep{pulse2024} teaches MLLMs to read rendered ECG images, and anyECG-chat~\citep{anyecgchat2025} handles flexible-length, multi-record input. Across these works a common recipe has largely converged: supervised fine-tuning followed by verifiable, rule-grounded rewards.

Yet almost every one of these systems operates on the same substrate: a single, short, fully-visible resting ECG, interpreted in one shot. Much of real cardiac monitoring does not fit this mold. Ambulatory and telemetry recordings span hours to days, and their most clinically decisive findings are \emph{paroxysmal}: a brief run of atrial fibrillation, a handful of ventricular ectopic beats, a multi-second pause, or a transient ST deviation, all buried in an otherwise unremarkable trace. A twenty-four-hour recording rendered at diagnostic resolution is not one image but thousands of pages. No fixed-size context can hold it, and no single glance can interpret it. Such a recording is also not analyzed in one batch after the fact. Telemetry is read as it streams in, and the clinical value of a finding degrades with how late it is reported.

This regime is \emph{structurally} agentic, and the reason is time's arrow rather than an imposed budget. Because the recording arrives online, an agent, human or model, at any moment sees only what has already elapsed. The future has not happened yet and is physically unobservable, while the past streams by and, unless deliberately committed to memory, is gone. This causal structure makes three competencies necessary rather than decorative, the same triad of memory, tool use, and planning that the agent literature has converged on~\citep{xi2025rise,wang2024survey}. \textbf{Memory}, because elapsed signal is expensive or impossible to revisit, so an agent must carry evidence forward and an event not recorded when it passes is effectively lost. \textbf{Tool use}, because with no pixels to eyeball the agent reads the waveform only through signal-grounded measurements of rhythm, ST level, and beat morphology, turning raw signal into quantities it can reason over. \textbf{Planning}, because with time continuously advancing the agent must decide what to measure now and at what granularity, coarsely over a long quiet stretch and finely over a suspicious one, and when to commit. None of this is an imposed cap on how much may be viewed; it is the streaming structure of the signal itself, which also removes a recurring shortcut in agentic benchmarks built on fully-observable inputs~\citep{liu2024agentbench,zhou2024webarena}, where a capable model can bypass the intended behaviour.

We introduce \textbf{ECG-Scroll}, which turns this setting into a clean, objective testbed with two parts, a benchmark and an agent environment. The benchmark streams real ambulatory recordings to an agent chunk by chunk across a suite of tasks spanning AF episode detection and burden, ischemia, ventricular ectopy, rare-event search, and change detection over five databases~\citep{mitbih2001,ltafdb,europeanstt1992,icentia11k2019}. ECG is unusually well suited to objective scoring: unlike the free-text impressions behind existing ECG question-answering benchmarks~\citep{ecgqa2023}, its ground truth is time-stamped and exactly checkable, since episode onsets and offsets, beat-level labels, and interval measurements are all machine-verifiable, so rewards are rule-based rather than judge-based. Retaining the signal also lets us check not just what the agent reports but \emph{when}, scoring the \textbf{detection latency} between an event's true onset and the moment it is first recorded. The agent environment is the fixed interaction layer any policy plugs into, a small set of causal actions over the elapsed signal and two tool families, signal tools that return numbers only and diagnostic tools that return a ready-made label. Together they make memory, tool use, and planning both necessary and measurable.

Our contributions are three:
\begin{itemize}
\item The \textbf{ECG-Scroll benchmark}, a long-horizon, streaming, causal formulation of ECG interpretation, instantiated as a diverse task suite over five databases, each with a rule-based verifier and a streaming-only detection-latency metric, and released as \nInstances{} whole-recording instances spanning \nHours{} hours.
\item The \textbf{ECG-Scroll agent environment}, a fixed, gym-style interaction layer of causal actions (\textsc{skim}, \textsc{measure}, \textsc{write}) and two tool families, clinician-ordered signal tools and diagnostic tools, that make memory, tool use, and planning necessary and measurable.
\item An \textbf{empirical study} that evaluates a signal-threshold rule agent and off-the-shelf LLM agents online, with and without the diagnostic tools, locating the benchmark's head-room.
\end{itemize}

\section{Related Work}
\subsection{ECG Multimodal LLMs and Reasoning}
Representation learning that aligns ECG waveforms with clinical text~\citep{merl2024,fgclep2025,liu2025two,anyecg2024,li2024biased} established that signals and reports can share a clinically meaningful space, and recent multimodal LLMs build on this substrate to interpret ECG directly. GEM~\citep{gem2025} links diagnoses to measurable parameters through a dual-encoder over an ECG time series, image, and text. ECG-R1~\citep{ecgr12026} trains on a protocol-guided corpus with diagnostic-evidence reinforcement learning and documents how widespread hallucination is across ECG MLLMs. PULSE~\citep{pulse2024} instruction-tunes on rendered ECG images, and anyECG-chat~\citep{anyecgchat2025} handles flexible-length, multi-record input. These systems show that verifiable-reward training and evidence grounding work well for ECG, but they all read short, fully-visible resting ECGs in a single forward pass: none model a recording that arrives over time, expose measurement as callable actions, or evaluate online planning and report timeliness. ECG-Scroll is complementary, targeting the long-horizon, streaming regime they leave open.

\subsection{Agentic LLMs: Memory, Tools, and Planning}
The agent literature~\citep{xi2025rise,wang2024survey} commonly decomposes capability into planning~\citep{yao2022react,yao2023tree,shinn2023reflexion}, memory~\citep{packer2023memgpt,park2023generative}, and tool use~\citep{schick2023toolformer}, and evaluates it with tool-agent-user or long-horizon web and task benchmarks such as $\tau$-bench~\citep{taubench2024}, AgentBench~\citep{liu2024agentbench}, WebArena~\citep{zhou2024webarena}, and ToolLLM~\citep{qin2024toolllm}. In parallel, R1-style reinforcement learning with verifiable rewards~\citep{shao2024deepseekmath,guo2025deepseek} has proven effective when correctness is machine-checkable. ECG-Scroll instantiates this triad in a clinically grounded domain where the causal structure of a streaming recording, not an imposed budget, makes the three competencies unavoidable: the past must be remembered rather than re-read, and rewards, including \emph{when} an event is reported, are objectively verifiable rather than model-judged.

\section{The ECG-Scroll Benchmark}
\label{sec:benchmark-note}

Existing ECG resources do not cover this setting. Model-facing corpora and QA benchmarks are built on short, fully-visible resting ECGs, while the long-duration ambulatory recordings that carry fine-grained temporal annotation sit in a mature but separate ecosystem that has not been used to evaluate MLLMs or agents. ECG-Scroll bridges the two, repurposing episode- and beat-level annotations from real ambulatory recordings as objective, time-stamped ground truth for online reading.

\begin{figure}[t]
\centering
\includegraphics[width=0.9\linewidth]{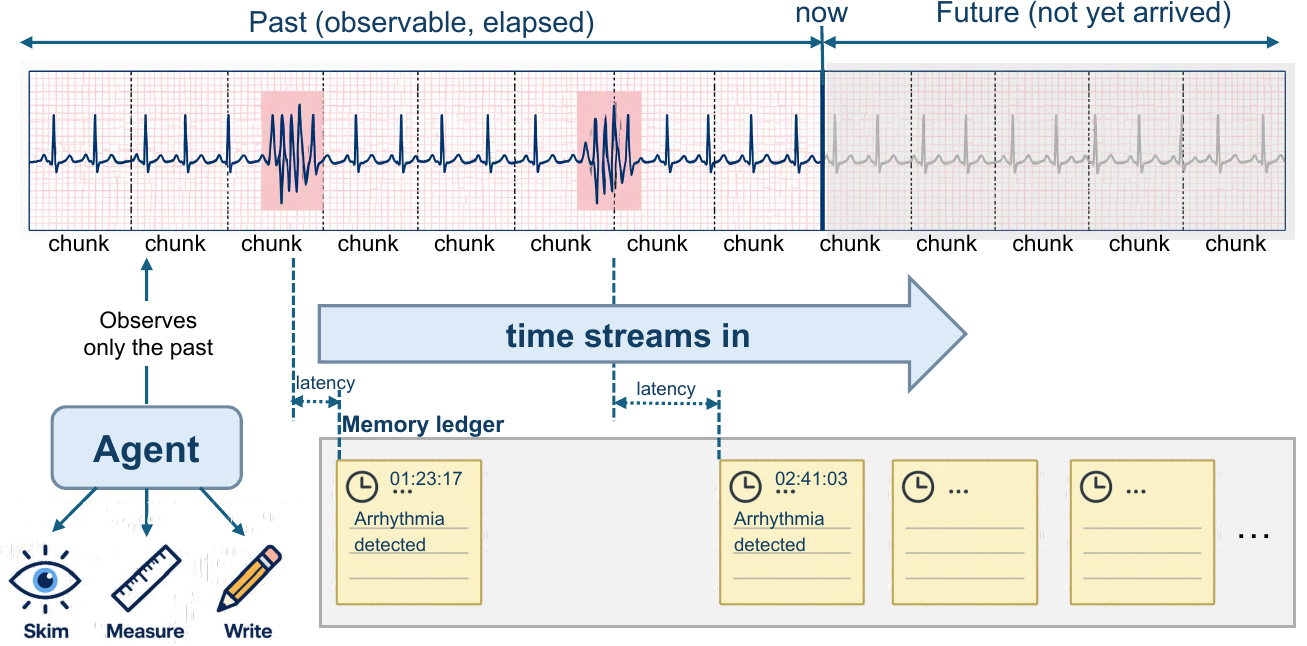}
\caption{ECG-Scroll at a glance. A long ambulatory recording streams left to right, and a monotonic cursor (\emph{now}) advances one chunk at a time, splitting it into the observable past and the not-yet-arrived future. The agent observes only the past: it measures the elapsed signal with signal-grounded tools and writes cursor-timestamped findings to memory.}
\label{fig:overview}
\end{figure}

\subsection{Problem Formulation}
A problem instance is a recording $R$ over a time span $[0,T]$, with $T$ from tens of minutes to a day, together with a hidden event timeline $E=\{(c_j,\ell_j,[a_j,b_j])\}_j$, where each event has a clinical type $c_j$ such as atrial fibrillation, ventricular ectopy, pause, or ST deviation, a lead $\ell_j$, and an onset/offset interval $[a_j,b_j]$. The recording is delivered \emph{online}: it is retained on the server and revealed to the agent chunk by chunk, never handed over in full, as Figure~\ref{fig:overview} illustrates.

We model interpretation as an \textbf{online, causal decision process}. Time advances through a monotonically increasing cursor $\kappa_t$; at step $t$ the agent may skim and measure the signal only over the elapsed span $[0,\kappa_t]$, while the future $(\kappa_t,T]$ has not arrived and is unobservable. It holds a structured memory $M_t$, its observation is the elapsed signal up to $\kappa_t$ together with the last tool result and $M_t$, and the event timeline $E$ is never observed directly. The action set is
\[
\mathcal{A}=\underbrace{\{\textsc{skim}(d)\}}_{\text{let time pass}}
\ \cup\ \{\textsc{measure}(\mathit{tool},t_0,t_1)\}\ \cup\ \{\textsc{write}(e)\},
\]
where $\textsc{skim}(d)$ advances the cursor by a duration $d$ the agent \emph{chooses}, coarsely over a long quiet stretch and finely over a suspicious one, and returns a lossy profile of that newly-elapsed span; $\textsc{measure}(\mathit{tool},t_0,t_1)$ runs one tool the agent \emph{chooses} over an elapsed sub-window without advancing time; and $\textsc{write}$ appends a cursor-timestamped entry to memory, from which the final answer is assembled. Choosing the skim granularity is the core planning lever. Re-measuring elapsed signal before the current skim window is permitted but costed as a rewind, so an agent that fails to commit passing evidence to $M_t$ pays for it later, and a strict variant forbids rewind entirely, leaving memory as the only access to the past. An episode ends when the stream is exhausted. The crux is causal, not budgetary: the future is never visible and the past survives only in memory, so the agent must plan what to measure now, retain it, and report events as they pass rather than after seeing everything.

Because every $\textsc{write}$ is timestamped, we additionally score \textbf{detection latency}: for each ground-truth event, the delay between its onset and the first memory entry that matches it at temporal $\mathrm{IoU}\ge\tau$. Events never recorded count as missed. Latency is a purely rule-based signal that batch evaluation cannot produce, and it directly rewards memory and planning.

\begin{figure}[t]
\centering
\includegraphics[width=0.9\linewidth]{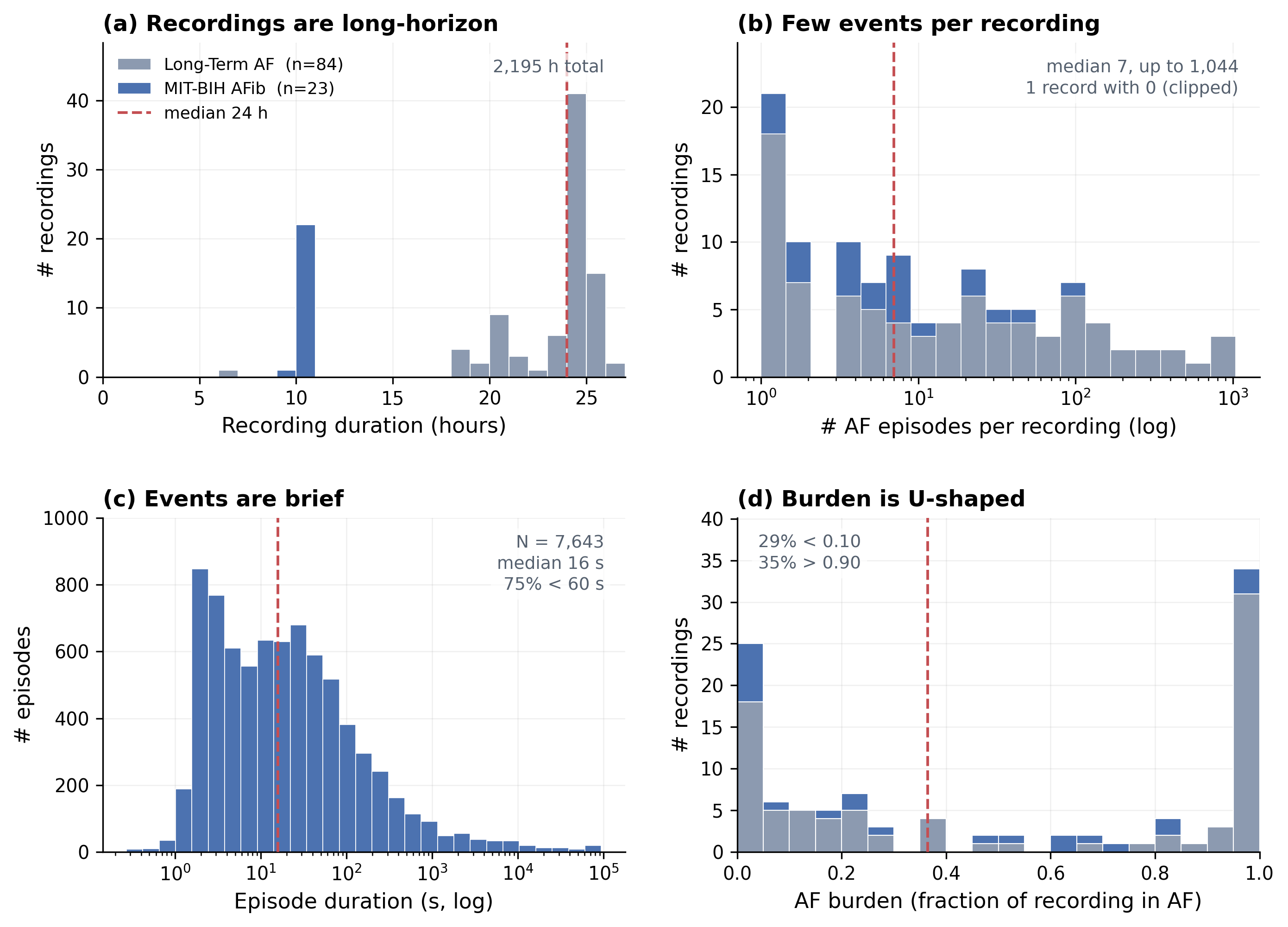}
\caption{The AF core. \emph{(a)} Whole-recording durations. \emph{(b)} AF episodes per recording (log axis). \emph{(c)} Individual AF episode durations (log axis). \emph{(d)} Per-recording AF burden, the fraction of time in AF.}
\label{fig:stats1}
\end{figure}

\subsection{Data and Streaming}
ECG-Scroll uses real two-lead ambulatory recordings from PhysioNet~\citep{goldberger2000physiobank}, the regime in which long-horizon online reading is clinically native and fine-grained temporal ground truth exists: the MIT-BIH Arrhythmia~\citep{mitbih2001}, MIT-BIH Atrial Fibrillation, and Long-Term AF~\citep{ltafdb} databases, the European ST-T Database~\citep{europeanstt1992}, and the large single-lead Icentia11K corpus~\citep{icentia11k2019}. Each recording is streamed to the agent in fixed-length chunks, 30\,s by default, in temporal order, and tools read the retained signal for the elapsed span only through the WFDB toolkit~\citep{sharma2023wfdb}. Because our agents read the waveform through signal-grounded measurement tools rather than pixels, no rendering is required, and retaining each signal lets tools and verifiers read exact values and timings.

The release is a suite of \nInstances{} whole-recording instances over \nRecordings{} recordings spanning \nHours{}~hours across five databases, each instance the entire recording over $[0,T]$ with no windowing, matching an online formulation whose only budget is time itself. Its core is atrial fibrillation, \nAFcore{} instances from all 23 MIT-BIH AFib and all 84 Long-Term AF recordings, each instantiated as an episode-detection instance and an AF-burden instance. These AF recordings run for hours, their episodes are few and brief with a median duration near 16\,s, and their per-recording burden is strongly U-shaped, so a verifier must span the full base-rate range rather than reward a single default guess. Figure~\ref{fig:stats1} makes this structure concrete, showing the whole-recording durations, the sparse and short episodes, and the U-shaped burden. The suite further spans ischemia on the European ST-T Database and ventricular ectopy on MIT-BIH Arrhythmia and Long-Term ST, together with rare-event search and change detection, broadening the event types an online reader must catch. Appendix~\ref{app:stats} details the task/database distribution and how little of each stream the target events occupy.

\subsection{Task Suite and Verifiers}
\begin{table}[t]
\centering
\caption{The ECG-Scroll task suite. Each task is a whole-recording streaming instance with an objective, rule-based verifier. $N$ is the number of whole-recording instances and Mean dur.\ their average length.}
\label{tab:tasks}
\small
\begin{tabular}{@{}llllrr@{}}
\toprule
Task & Source & Answer & Verifier / Metric ($\uparrow$) & $N$ & Mean dur. \\
\midrule
Episode detection & afdb, ltafdb & interval set & temporal $F_1$ (IoU$\ge\tau$) & 107 & 20.5\,h \\
AF burden & afdb, ltafdb & scalar & rel.\ error score & 107 & 20.5\,h \\
Change detection & afdb & change-point & hit@$\pm10$s ($|\hat t-t^\star|\le\delta$) & 23 & 10.2\,h \\
Ischemia detection & edb & interval set & temporal $F_1$ (IoU$\ge0.75$) & 85 & 2.0\,h \\
PVC burden & mitdb, ltdb & scalar & rel.\ error score & 55 & 3.1\,h \\
Rare-event search & mitdb, ltdb & interval & hit@$\pm15$s (midpoint tol.) & 13 & 0.5\,h \\
\bottomrule
\end{tabular}
\end{table}
All five tasks admit an objective, rule-based verifier and use no model-as-judge, and Table~\ref{tab:tasks} lists them with their sources and metrics. Episode and ischemia detection are scored by temporal $F_1$ over interval matches at $\mathrm{IoU}\ge\tau$, burden by a relative-error score, and change detection and rare-event search by a localization tolerance. Appendix~\ref{app:verifiers} gives the exact definition of each verifier.

\section{The ECG-Scroll Agent}
\label{sec:agent}
The benchmark specifies what is evaluated. This section specifies how an agent interacts with a streaming recording: the action interface, the clinically-motivated measurement tools, and the memory. This layer is a fixed, gym-style environment that any policy plugs into, from the rule reference and LLM agents we study to a future trained model, so it is a contribution in its own right, separate from the benchmark it instruments.

\subsection{Actions}
At each step the agent issues one action from $\mathcal{A}=\{\textsc{skim},\textsc{measure},\textsc{write}\}$, all computed from the retained signal up to the current cursor $\kappa_t$ and never the future. Table~\ref{tab:tools} collects the full interface, the control actions together with the signal and diagnostic tools the model chooses among when it calls \textsc{measure}. $\textsc{skim}(d)$ advances the cursor by a duration $d$ the agent chooses and returns a lossy profile of the newly-elapsed span, a peak feature and where it lies, so a short buried event surfaces as an elevated peak to examine; choosing the granularity, coarse to cover ground and fine to localize, is the core planning lever. $\textsc{measure}(\mathit{tool},t_0,t_1)$ is a broad action that runs one tool the model names over an elapsed sub-window without advancing time, choosing among the signal and diagnostic tools below; re-measuring before the current skim window is a costed rewind, or is forbidden in the strict variant, so the past is reliably available only through what was written down. $\textsc{write}$ commits a cursor-timestamped finding to memory, from which the answer is assembled when the stream ends.

\subsection{Signal Tools: a Clinician's Reading Order}
\label{sec:signaltools}
The agent never sees pixels; the tools' numeric output is its only observation. Rather than an ad hoc set matched to the current tasks, we design the tools the way a cardiologist reads a strip, systematically band by band: rate and rhythm, then the P wave, the QRS complex, the ST segment, the T wave, and the interval measurements. This ordering mirrors clinical practice, makes the tool set complete with respect to the diagnostic content of a two-lead strip, and lets ECG-Scroll extend beyond the tasks released here, to atrioventricular block, bundle-branch block and hypertrophy, long or short QT and drug effects, hyperkalemia, and atrial flutter or pauses, without adding new primitives. Every signal tool returns numbers only, a heart rate, an RR coefficient of variation, an ST deviation in millivolts, a QRS width in milliseconds, all computed from beats located by a Pan-Tompkins detector~\citep{pan1985real}, and none returns a diagnostic label, so the agent must reason from measurements to a verifiable conclusion, exactly as a reader reasons from calipers to an interpretation. Each tool reports a compact summary by default and the underlying per-beat arrays on request.

\begin{table}[t]
\centering
\caption{The ECG-Scroll agent's actions and tools. Control actions drive the causal stream. Signal tools, which the model chooses among when it calls \texttt{measure}, are organized in a clinician's reading order and return numbers only. Diagnostic tools instead return a ready-made classification; we evaluate agents with and without them.}
\label{tab:tools}
\small
\begin{tabular}{@{}p{2.9cm}p{2.0cm}p{6.3cm}@{}}
\toprule
Action / tool & Target & Output / effect \\
\midrule
\multicolumn{3}{@{}l}{\emph{Control actions} (drive the causal stream)}\\
\texttt{skim(duration)} & cover / plan & advance the cursor by \texttt{duration}\,s; return the newly-elapsed span's lossy profile (peak feature + coarse sub-bins) \\
\texttt{measure(tool,t0,t1)} & localize & run one tool the model \emph{chooses} over an elapsed sub-window without advancing time; picks among the signal (and, under \texttt{+dx}, diagnostic) tools below \\
\texttt{write(entry)} & commit & append a cursor-timestamped finding to memory; the answer is assembled from these when the stream ends \\
\midrule
\multicolumn{3}{@{}l}{\emph{Signal tools} (numbers only, no label; clinician's reading order)}\\
\texttt{get\_rate\_rhythm} & rate, AF, pauses & hr\_bpm, rr\_mean/sd\_ms, rr\_irregularity (CV), rr\_min/max (\emph{per-beat:} rr\_intervals\_ms) \\
\texttt{get\_p\_wave} & flutter, conduction & p\_present\_frac, pr\_mean\_ms, p\_amp, pp\_regularity (\emph{per-beat:} pr\_intervals\_ms) \\
\texttt{get\_qrs} & ectopy, BBB, LVH & mean\_qrs\_ms, wide\_beat\_frac, qrs\_amp, ectopic\_count/times (\emph{per-beat:} qrs\_widths\_ms) \\
\texttt{get\_st} & ischemia & st\_mv per lead (J+80\,ms vs.\ PR baseline), sign (elevation/depression) (\emph{per-beat:} ST samples) \\
\texttt{get\_t\_wave} & ischemia, K$^+$ & t\_amp per lead, polarity (\emph{per-beat:} T amplitudes) \\
\texttt{get\_qt} & long/short QT & qt\_mean\_ms, qtc\_ms (Bazett) (\emph{per-beat:} qt\_intervals\_ms) \\
\midrule
\multicolumn{3}{@{}l}{\emph{Diagnostic tools} (return a ready-made label; a with/without axis)}\\
\texttt{classify\_rhythm} & rhythm label & dominant\_rhythm (sinus/AF/\ldots), af\_frac \\
\texttt{classify\_beats} & beat labels & per-beat types (N/V/E/\ldots), pvc\_times \\
\texttt{flag\_ischemia} & ischemia label & ischemic (bool), lead, st\_mv, episodes \\
\bottomrule
\end{tabular}
\end{table}

\subsection{Diagnostic Tools}
Alongside the signal tools we provide diagnostic tools that return a ready-made classification rather than a measurement: a rhythm classifier, a per-beat classifier, and an ischemia flag. Where a signal tool reports an RR coefficient of variation and leaves the agent to conclude atrial fibrillation, a diagnostic tool reports that conclusion directly. Their outputs are causally clamped to the elapsed span like every other tool. We evaluate each agent both with and without the diagnostic tools: the without setting keeps the agent reasoning from measurements alone, and the comparison isolates whether a ready-made label actually improves online localization and latency or is merely leaned on in place of signal reasoning.

\subsection{Memory}
Memory is a typed findings ledger: each $\textsc{write}$ appends an entry of type, lead, interval, value, and confidence stamped with the current cursor time, and the current ledger is always part of the observation. This gives a clean, inspectable substitute for an unbounded free-text scratchpad, makes memory content directly scorable, and, because entries are timestamped, makes detection latency measurable. In the streaming setting memory is not optional bookkeeping: since elapsed signal is expensive or impossible to revisit, an event not written down when it passes is effectively lost.

\section{Experiments}
\label{sec:reference}

\begin{table}[t]
\centering
\caption{Planning agent: the model drives its own scan. Each model is run without (\texttt{--dx}) and with (\texttt{+dx}) the diagnostic tools, against the Rule Agent reference. Metrics are accuracies in $[0,1]$, higher is better.}
\label{tab:main-plan}
\small
\setlength{\tabcolsep}{4pt}
\resizebox{0.85\textwidth}{!}{%
\begin{tabular}{@{}ll cccc cc@{}}
\toprule
& & Episode & AF burden & Change & Ischemia & PVC burden & Rare-event \\
Agent & Tier & ($F_1\uparrow$) & (score$\uparrow$) & (hit@$\pm$10s$\uparrow$) & ($F_1\uparrow$) & (score$\uparrow$) & (hit@$\pm$15s$\uparrow$) \\
\midrule
Rule Agent  & --dx & 0.02 & 0.26 & 0.17 & 0.07 & 0.34 & 0.00 \\
            & +dx  & 0.52 & 0.83 & 0.39 & 0.80 & 0.25 & 0.85 \\
\midrule
Qwen3-8B      & --dx & 0.03 & 0.17 & 0.13 & 0.00 & 0.47 & 0.00 \\
              & +dx  & 0.03 & 0.24 & 0.17 & 0.00 & 0.58 & 0.46 \\
\cmidrule(lr){1-8}
Qwen3-30B     & --dx & 0.03 & 0.17 & 0.17 & 0.00 & 0.47 & 0.00 \\
              & +dx  & 0.03 & 0.24 & 0.17 & 0.00 & 0.57 & 0.46 \\
\cmidrule(lr){1-8}
Qwen3-235B    & --dx & 0.03 & 0.17 & 0.17 & 0.00 & 0.49 & 0.00 \\
              & +dx  & 0.03 & 0.24 & 0.17 & 0.00 & 0.57 & 0.31 \\
\cmidrule(lr){1-8}
Qwen3-30B-Thinking & --dx & 0.03 & 0.17 & 0.17 & 0.00 & 0.50 & 0.00 \\
              & +dx  & 0.03 & 0.21 & 0.17 & 0.00 & 0.51 & 0.15 \\
\cmidrule(lr){1-8}
GLM-4.5-Air   & --dx & 0.03 & 0.17 & 0.17 & 0.00 & 0.48 & 0.08 \\
              & +dx  & 0.03 & 0.24 & 0.17 & 0.00 & 0.58 & 0.46 \\
\cmidrule(lr){1-8}
Llama-3.3-70B & --dx & 0.03 & 0.18 & 0.17 & 0.03 & 0.49 & 0.00 \\
              & +dx  & 0.03 & 0.22 & 0.17 & 0.00 & 0.58 & 0.46 \\
\cmidrule(lr){1-8}
DeepSeek-V4.1 & --dx & 0.01 & 0.23 & 0.04 & 0.01 & 0.47 & 0.08 \\
\hphantom{DeepSeek-V4.1} & +dx  & 0.01 & 0.24 & 0.00 & 0.01 & 0.60 & 0.46 \\
\bottomrule
\end{tabular}}
\end{table}

Our study asks a single question: when a long recording streams past and its clinically decisive moments are brief and sparse, can an off-the-shelf language model read it online. The benchmark makes the answer measurable, and the agent environment makes it fair, since every policy plugs into the same causal actions, the same signal and diagnostic tools, and the same rule-based verifiers. We are not building a state-of-the-art detector; we want to locate which of the three streaming competencies a capable model already has and which it still lacks.

To separate those competencies we evaluate the same models in two configurations that differ in one respect, who decides where to look. The Planning agent drives the cursor itself, choosing how coarsely to skim, when to stop and measure, and when to commit a finding, so it exercises planning, memory, and tool use together. The Judge agent keeps the memory and the tools but removes the planning: the environment walks the record chunk by chunk and the model only judges which of the elapsed chunks contain an event. Reading the two side by side isolates planning from everything else, because any gap between them cannot come from perception or knowledge, which are held fixed, but only from the act of self-directed scanning that one configuration has and the other does not. Throughout, \texttt{--dx} denotes the signal tools alone and \texttt{+dx} adds the diagnostic tools that return a ready label, and we run every agent both ways to see whether a ready verdict actually helps online.

\paragraph{Setup.} Each whole recording is streamed online and scored by the task verifier at the end of the stream. For the Judge agent the environment presents the elapsed signal in fixed thirty-second chunks, grouped into time-ordered batches, and the model flags the ones it believes are events; for the Planning agent the model instead issues its own actions over the elapsed signal and chooses how far to advance the cursor at each step. In both cases flagged or written spans are committed to the memory ledger with their cursor timestamp and assembled into the task answer by the same rules, so the two configurations are scored identically. As a signal-grounded reference we include a Rule Agent that uses no language model: walking the stream in the same fixed chunks, it measures the elapsed slice and flags a chunk when the relevant cue crosses a threshold under \texttt{--dx}, or when the diagnostic label fires under \texttt{+dx}. It is a strong measurement-plus-threshold baseline rather than an upper bound, and the language models exceed it on a few cells. We evaluate open-weight models across scales and families, Qwen3~\citep{yang2025qwen3} at 8B, 30B, and 235B with a reasoning-tuned variant, GLM~\citep{zeng2025glm}, and Llama~\citep{grattafiori2024llama}, together with the proprietary DeepSeek-V4.1-Flash~\citep{xu2026deepseek} queried through its API. No model is trained on the benchmark.

\begin{table}[t]
\centering
\caption{Judge agent: the environment walks the record chunk by chunk and the model only judges which elapsed chunks are events, with planning removed.}
\label{tab:main-judge}
\small
\setlength{\tabcolsep}{4pt}
\resizebox{0.85\textwidth}{!}{%
\begin{tabular}{@{}ll cccc cc@{}}
\toprule
& & Episode & AF burden & Change & Ischemia & PVC burden & Rare-event \\
Agent & Tier & ($F_1\uparrow$) & (score$\uparrow$) & (hit@$\pm$10s$\uparrow$) & ($F_1\uparrow$) & (score$\uparrow$) & (hit@$\pm$15s$\uparrow$) \\
\midrule
Rule Agent  & --dx & 0.02 & 0.26 & 0.17 & 0.07 & 0.34 & 0.00 \\
            & +dx  & 0.52 & 0.83 & 0.39 & 0.80 & 0.25 & 0.85 \\
\midrule
Qwen3-8B      & --dx & 0.02 & 0.30 & 0.09 & 0.04 & 0.23 & 0.00 \\
              & +dx  & 0.50 & 0.96 & 0.35 & 0.80 & 0.45 & 0.77 \\
\cmidrule(lr){1-8}
Qwen3-30B     & --dx & 0.01 & 0.31 & 0.04 & 0.05 & 0.41 & 0.00 \\
              & +dx  & 0.51 & 0.96 & 0.39 & 0.80 & 0.38 & 0.85 \\
\cmidrule(lr){1-8}
Qwen3-235B    & --dx & 0.02 & 0.31 & 0.09 & 0.10 & 0.46 & 0.00 \\
              & +dx  & 0.52 & 0.99 & 0.39 & 0.80 & 0.47 & 0.85 \\
\cmidrule(lr){1-8}
Qwen3-30B-Thinking & --dx & 0.02 & 0.31 & 0.04 & 0.05 & 0.38 & 0.00 \\
              & +dx  & 0.50 & 0.97 & 0.39 & 0.80 & 0.38 & 0.77 \\
\cmidrule(lr){1-8}
GLM-4.5-Air   & --dx & 0.01 & 0.31 & 0.17 & 0.05 & 0.18 & 0.00 \\
              & +dx  & 0.29 & 0.57 & 0.17 & 0.54 & 0.46 & 0.85 \\
\cmidrule(lr){1-8}
Llama-3.3-70B & --dx & 0.02 & 0.32 & 0.17 & 0.06 & 0.22 & 0.00 \\
              & +dx  & 0.47 & 0.98 & 0.35 & 0.80 & 0.47 & 0.85 \\
\cmidrule(lr){1-8}
DeepSeek-V4.1 & --dx & 0.03 & 0.32 & 0.13 & 0.12 & 0.37 & 0.00 \\
\hphantom{DeepSeek-V4.1} & +dx  & 0.52 & 0.99 & 0.17 & 0.80 & 0.51 & 0.92 \\
\bottomrule
\end{tabular}}
\end{table}

\paragraph{Planning is the missing competency.} When the model has to plan its own scan, it fails on the tasks that demand sustained coverage of a long recording, and as Table~\ref{tab:main-plan} shows, the diagnostic label does not rescue it there. Episode $F_1$ sits near the floor, ischemia collapses to zero, and change detection and AF burden stay low whether or not the ready label is available, and the pattern holds from 8B to 235B and for the reasoning-tuned model alike, so neither scale nor extra test-time reasoning changes it. A ready verdict is worth nothing on these tasks if the agent never turns it into a committed finding, which is exactly what happens: the Rule Agent, given the same tools, recovers strong accuracy under \texttt{+dx}, so the information needed to succeed is plainly present in the environment; the Planning agent simply fails to act on it over tens of thousands of seconds. The exception proves the rule. On the two tasks whose targets are point-like and whose records are short, PVC burden and rare-event search, the Planning agent does climb well off the floor once it can choose the right tool, reaching PVC burden around $0.5$--$0.6$ and rare-event search up to $0.46$ under \texttt{+dx} across models; here a handful of tight commits suffices and the closed loop is short enough to run. The binding constraint is therefore long-horizon planning: where success requires covering a multi-hour stream and committing many findings, the model defers or floods and the label cannot help; where a single well-placed commit will do, it can act on the same label.

\paragraph{Judgment is not.} Removing that one degree of freedom changes the picture completely. The Judge results in Table~\ref{tab:main-judge} tell a different story: when the environment handles coverage and the model only decides which elapsed chunks are events, the same models rise to essentially match the signal-grounded reference once the diagnostic label is available, recovering AF burden and episode detection and reaching the reference on ischemia and rare-event search. The competency was there all along: these models can recognize an event in a chunk placed in front of them and commit it in time, and they lose that ability only when they must also decide, on their own, when to look and when to stop. Two further patterns stand out. Without the diagnostic label the Judge agent still collapses and floods nearly every chunk as an event, so the gain reflects genuine use of the label rather than the removal of planning alone. And scale again barely moves the numbers, with one clean exception in PVC burden, where accuracy grows with model size because that task rewards reading the raw waveform rather than trusting a label.

\begin{figure}[t]
\centering
\includegraphics[width=\linewidth]{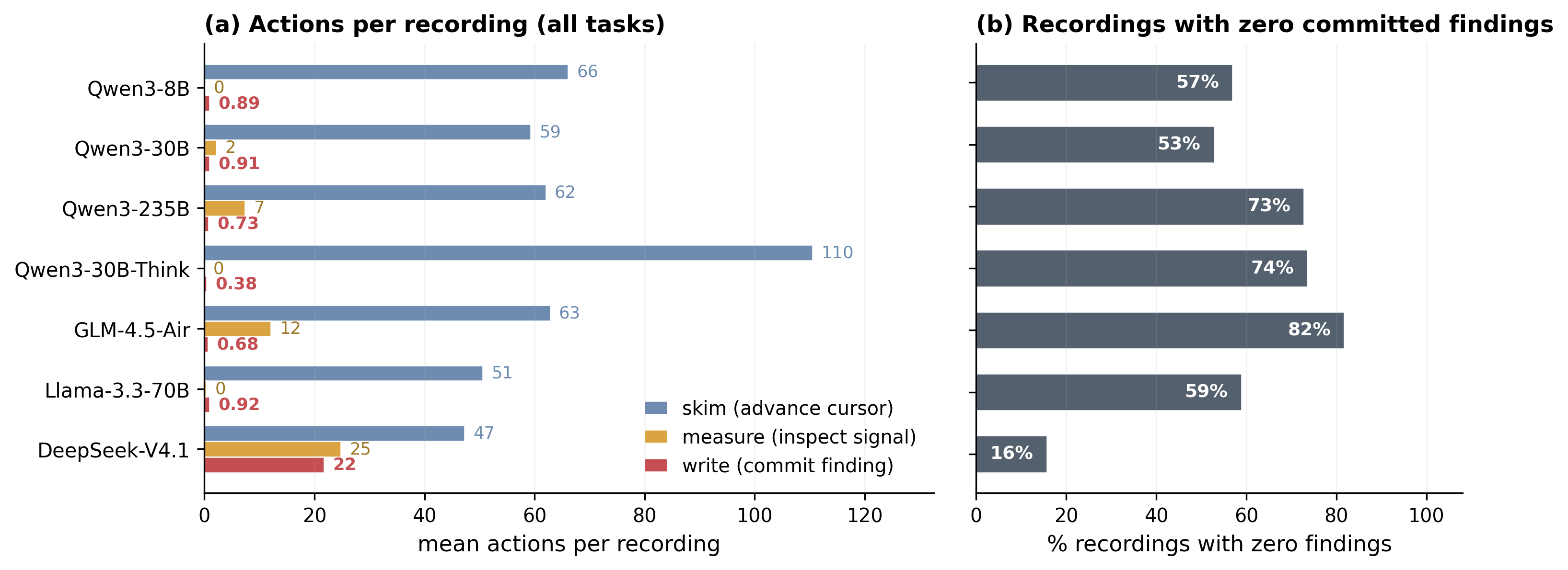}
\caption{Planning agent behavior aggregated across all six tasks with the diagnostic tools (task-equal-weighted). Left: mean skim, measure, and write actions per recording. Right: fraction of recordings ending with no committed finding.}
\label{fig:whyplanning}
\end{figure}

\paragraph{Where planning breaks down.} Figure~\ref{fig:whyplanning} summarizes what the Planning agent actually does across all six tasks, and the failure is not one of looking. Every model advances the cursor freely, tens to over a hundred skims per recording, and some measure heavily on top of that, GLM and the 235B model especially. What almost never follows is a useful commit: most models write close to nothing, ending half to two-thirds of recordings with an empty ledger, and the ones that measure the most commit the least, so the signal reaches the model and is never recorded. DeepSeek is the lone exception, and only on the long atrial-fibrillation recordings, where it fails the other way, writing on nearly every recording but so loosely that almost nothing survives the overlap criterion; on the shorter, sparser ischemia, ectopy, and rare-event tasks it under-commits like the rest. Under- or over-committing, the same thing is missing: a closed loop that decides when a suspicion is confirmed and pins it down precisely enough to count. Handing that decision to the environment lifts the same models from the floor to near-reference accuracy, so autonomous stop-and-commit control, not measurement or recognition, is the competency they most clearly lack. One consequence is that detection latency cannot yet be read off meaningfully: too few well-localized commits leave too few matched events to time. It is not a limitation of the benchmark but a measurement awaiting agents strong enough to be worth timing, and once a policy commits reliably it becomes a central axis on which they will be judged.

\section{Conclusion}
We introduced ECG-Scroll, which recasts long-recording ECG interpretation as an online, causal decision process and instantiates it as both a benchmark with objective verifiers and a fixed agent environment where memory, tool use, and planning are necessary. Evaluating models with and without control over their own scan reveals a sharp asymmetry: they can recognize and commit an event placed in front of them, yet cannot decide on their own when to look and when to write, and this open-loop planning collapses across scales and reasoning effort. Autonomous stop-and-commit control over a streaming signal is the skill long-horizon agents most lack, and we release the benchmark and environment to make progress on it measurable.

\subsection*{AI use statement}
We used generative AI tools to polish the writing of this paper, improving its grammar and readability.

\bibliographystyle{iclr2027_conference}
\bibliography{refs}

\appendix
\section{Additional Dataset Statistics}
\label{app:stats}
\begin{figure}[h]
\centering
\includegraphics[width=\linewidth]{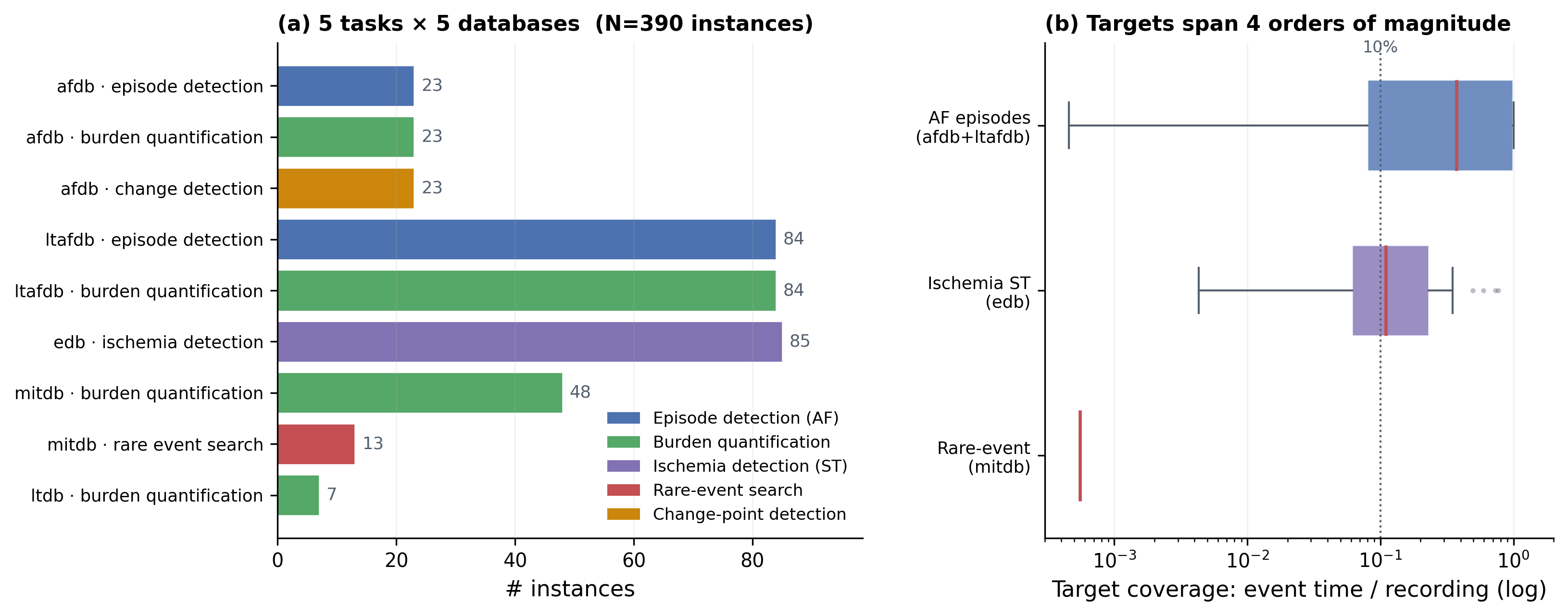}
\caption{Task and dataset coverage. \emph{(a)} Instance counts across the five task types and five databases. \emph{(b)} Fraction of each stream occupied by target events, from AF episodes and transient ischemic ST-episodes down to near-instantaneous rare-event targets (log axis).}
\label{fig:stats_multi}
\end{figure}

Figure~\ref{fig:stats_multi} details how the released instances distribute across the five task types and five databases, and how small a fraction of each stream the target events occupy.

\section{Task Verifiers}
\label{app:verifiers}
Each task is scored by an objective, rule-based verifier. We match a predicted interval $\hat{I}$ to a ground-truth interval $I$ when their temporal intersection-over-union satisfies $\mathrm{IoU}(\hat{I},I)\ge\tau$, with $\tau{=}0.5$ by default.
\begin{itemize}
\item \textbf{Episode detection \& localization.} Output a set of episode intervals for a given rhythm. Score is temporal $F_1$ over IoU-matched episodes (afdb, ltafdb).
\item \textbf{Burden quantification.} Output a scalar $\hat{v}$ (e.g.\ AF burden $=\!$ time-in-AF$/T$, or ventricular-ectopic beat fraction). Score $=\max\!\big(0,\,1-|\hat{v}-v|/\max(v,v_0)\big)$ with a task floor $v_0$; evaluated separately as AF burden (afdb, ltafdb) and PVC burden (mitdb, ltdb).
\item \textbf{Change detection.} Output a change-point $\hat{t}$ (e.g.\ sinus$\to$AF onset); reward $\mathbb{1}[|\hat{t}-t^\star|\le\delta]$, $\delta{=}10$\,s (afdb).
\item \textbf{Ischemia detection.} Output a set of ischemic ST-episode intervals; score is temporal $F_1$ over IoU-matched episodes (edb). Because ST-episode boundaries are clinically meaningful, we require a stricter overlap here ($\tau{=}0.75$) than for AF episode detection. A richer localization variant additionally scores exact lead and ST-magnitude $|\Delta\mathrm{ST}|$ within $\epsilon$\,mV.
\item \textbf{Rare-event search.} Localize a single short event in a long trace; because the target is a near-instantaneous beat ($\sim$1\,s), an $\mathrm{IoU}$ criterion is unreachable at the streaming chunk resolution, so we score localization by tolerance, $\text{hit}=\mathbb{1}[\,|\hat{t}-t^\star|\le\delta\,]$ with $\delta{=}15$\,s (half a chunk), where $\hat{t}$ is the predicted interval's midpoint (mitdb, ltdb).
\end{itemize}

\section{Agent Prompts and Action Interface}
\label{app:prompts}
Both agents share the same models, tools, and verifiers and differ only in who plans the scan. Here we give the exact instructions each receives.

\paragraph{Planning agent.} The Planning agent controls the cursor and issues one tool call per turn from $\{\texttt{skim},\texttt{measure},\texttt{write},\texttt{finish}\}$. Here \texttt{measure} is a single broad action that takes a \texttt{tool} argument: the model names which tool to run, choosing among the six signal tools always, plus the three diagnostic tools under \texttt{+dx}. Its system prompt establishes the online reading setup, the tool menu, and the scan strategy it is expected to run:
\begin{promptbox}{System prompt (Planning agent)}
You are an expert cardiologist reading a long ambulatory ECG that streams in ONLINE, in time order. You cannot see the waveform or the future; you read it only by calling tools on the already-elapsed signal, and you must plan your own scan. Your core lever is \texttt{skim(duration)}: YOU choose how big the next slice is. Over a long quiet stretch, skim BIG (300--1800\,s) to cover ground fast; when something looks suspicious, skim SMALL (5--60\,s) to look closely. Each skim returns a lossy profile of the newly-elapsed span, a peak value and where it lies, so a short buried event shows up only as an elevated peak; when you see one, do not skim forward, instead skim or measure that range again at fine resolution to pin it, then record it. \texttt{measure(tool, t0, t1)} is a single action that runs ONE tool of YOUR choice over an elapsed sub-window, without advancing time. YOU pick which tool. The signal tools return numbers only: \texttt{get\_rate\_rhythm} (heart rate + RR irregularity, the AF cue), \texttt{get\_p\_wave}, \texttt{get\_qrs} (beat morphology / ectopy), \texttt{get\_st} (ST-segment level, the ischemia cue), \texttt{get\_t\_wave}, \texttt{get\_qt}. Choose the tool whose measurement answers the current task, then reason from the numbers to a conclusion. The strategy that works: skim big $\rightarrow$ peak elevated? $\rightarrow$ skim/measure that range small to localize $\rightarrow$ confirm it $\rightarrow$ extend the boundary before and after to find where the event starts and ends $\rightarrow$ \texttt{write(start,end)} the whole span $\rightarrow$ continue. An event you never write scores zero. When the cursor reaches total, call \texttt{finish()}.
\end{promptbox}
A short task-specific instruction follows this shared preamble, naming the cue and, under \texttt{+dx}, the diagnostic tool for that task (for AF, high \texttt{rr\_irregularity} via \texttt{get\_rate\_rhythm}, confirmed by \texttt{classify\_rhythm}). After each action the agent receives a compact observation line reporting the cursor position, the number of events written so far, and what the last tool returned, such as the peak value and its location from a skim or the tool used and its measurement; when it has confirmed an event but committed nothing, the line reminds it that an unrecorded event scores zero. The model chooses every duration, window, and tool itself.

\paragraph{Judge agent.} The Judge agent never controls the cursor. The environment walks the recording in fixed thirty-second chunks grouped into time-ordered batches, and for each batch the model makes a single \texttt{flag\_chunks(indices)} call naming the chunks it believes contain the event. Its system prompt is:
\begin{promptbox}{System prompt (Judge agent)}
You are an expert cardiologist reading a long ambulatory ECG that streams in ONLINE, in time order, one 30-second chunk at a time. You cannot see the future. You are localizing every AF EPISODE; flag every chunk that is in AF. For each chunk you are given a ready-made diagnostic LABEL for that 30\,s window (rhythm class and \texttt{af\_frac}); trust the label and flag the chunk when the label says it is in AF. You will see chunks in time-ordered batches. For each batch, call \texttt{flag\_chunks(indices)} with the chunk indices that belong to the event; if none qualify, call it with an empty list. Call \texttt{flag\_chunks} exactly once per batch.
\end{promptbox}
Under \texttt{--dx} each chunk line instead carries signal measurements for that window, such as RR variability, heart rate, ST level, or wide-beat fraction, and the model decides from the numbers. In both configurations the flagged or written spans are committed to the same memory ledger with their cursor timestamp and assembled into the task answer by the same rules, so the two are scored identically.

\section{Successful Planning Trajectories}
\label{app:success}
The Planning agent is not uniformly at the floor: on PVC burden, where the target beats are point-like and the records are short enough to scan finely, it runs the intended closed loop, coarse-skim the whole record, measure where a peak rises, write a tight interval around each confirmed ectopic beat, and move on. We give two DeepSeek-V4.1 trajectories on MIT-BIH records ($T\approx1806$\,s), abbreviating repeated quiet skims. Cursor times are in seconds; each \texttt{write} is a $\sim$2\,s interval around one beat, and the burden is the union of these over $T$.

\paragraph{With the diagnostic tool (\texttt{+dx}), record \texttt{mitdb-118} (score 0.96; predicted burden 0.0078 vs.\ true 0.0070).} The agent chooses \texttt{classify\_beats} at each elevated peak, reads the returned \texttt{pvc\_times}, and commits one tight interval per beat; where the label reports no ventricular beats it skims on without writing.
\begin{lstlisting}[style=traj]
skim 600            -> peak 0.3 @ [0,20]
measure(classify_beats, 0..120)     -> pvc_times=[30.7, 106.9];        write [29.7, 31.7]
skim 300; measure(classify_beats, 120..240)  -> pvc_times=[224.6];     write [223.6, 225.6]
skim 300; measure(classify_beats, 360..480)  -> pvc_times=[461.0];     write [460.0, 462.0]
skim 300; measure(classify_beats, 480..600)  -> pvc_times=[494.1, 500.8, 567.8];  write [493.1, 495.1]
skim 300; measure(classify_beats, 600..720)  -> pvc_times=[]           (skim on, no write)
skim 300; measure(classify_beats, 720..840)  -> pvc_times=[743.9];     write [742.9, 744.9]
...  skim / measure across the quiet middle  ...
measure(classify_beats, 1320..1440) -> pvc_times=[1358.9, 1383.1];     write [1357.9, 1359.9]
measure(classify_beats, 1560..1680) -> pvc_times=[1586.8, 1588.7];     write [1585.8, 1587.8]
skim to end -> peak 0.0;  finish()
\end{lstlisting}

\paragraph{Signal tools only (\texttt{--dx}), record \texttt{mitdb-202} (score 0.83; predicted burden 0.0055 vs.\ true 0.0089).} Without a ready label the agent reasons from \texttt{get\_qrs} numbers, treating an elevated \texttt{ectopic\_count} / \texttt{wide\_beat\_frac} as the cue and writing around the returned \texttt{ectopic\_times}.
\begin{lstlisting}[style=traj]
skim 600 x3         -> peak 0.0    (quiet; scan on)
skim 600; measure(get_qrs, 360..480) -> ectopic_count=1, wide_frac=0.009, times=[476.6];  write [475.6, 477.6]
skim 300; measure(get_qrs, 480..600) -> ectopic_count=2, times=[570.6, 582.6];            write [569.6, 571.6]
skim 300; measure(get_qrs, 720..840) -> ectopic_count=4, wide_frac=0.035;                 write [735.6, 737.6]
skim 300; measure(get_qrs, 840..960) -> ectopic_count=4, wide_frac=0.037;                 write [870.0, 872.0]
skim 300; measure(get_qrs, 960..1080)-> ectopic_count=1;                                  write [989.3, 991.3]
skim to end -> peak 0.0;  finish()
\end{lstlisting}
Both runs show the same competent pattern the harness rewards: adaptive granularity (big skims over quiet stretches, fine measurement at each peak), a correct tool choice for the task, and a tight commit per confirmed beat. The gap to the Rule Agent on this task is one of recall among the many beats rather than of the loop itself; the failures documented in Section~\ref{sec:reference} are where this loop never closes.

\end{document}